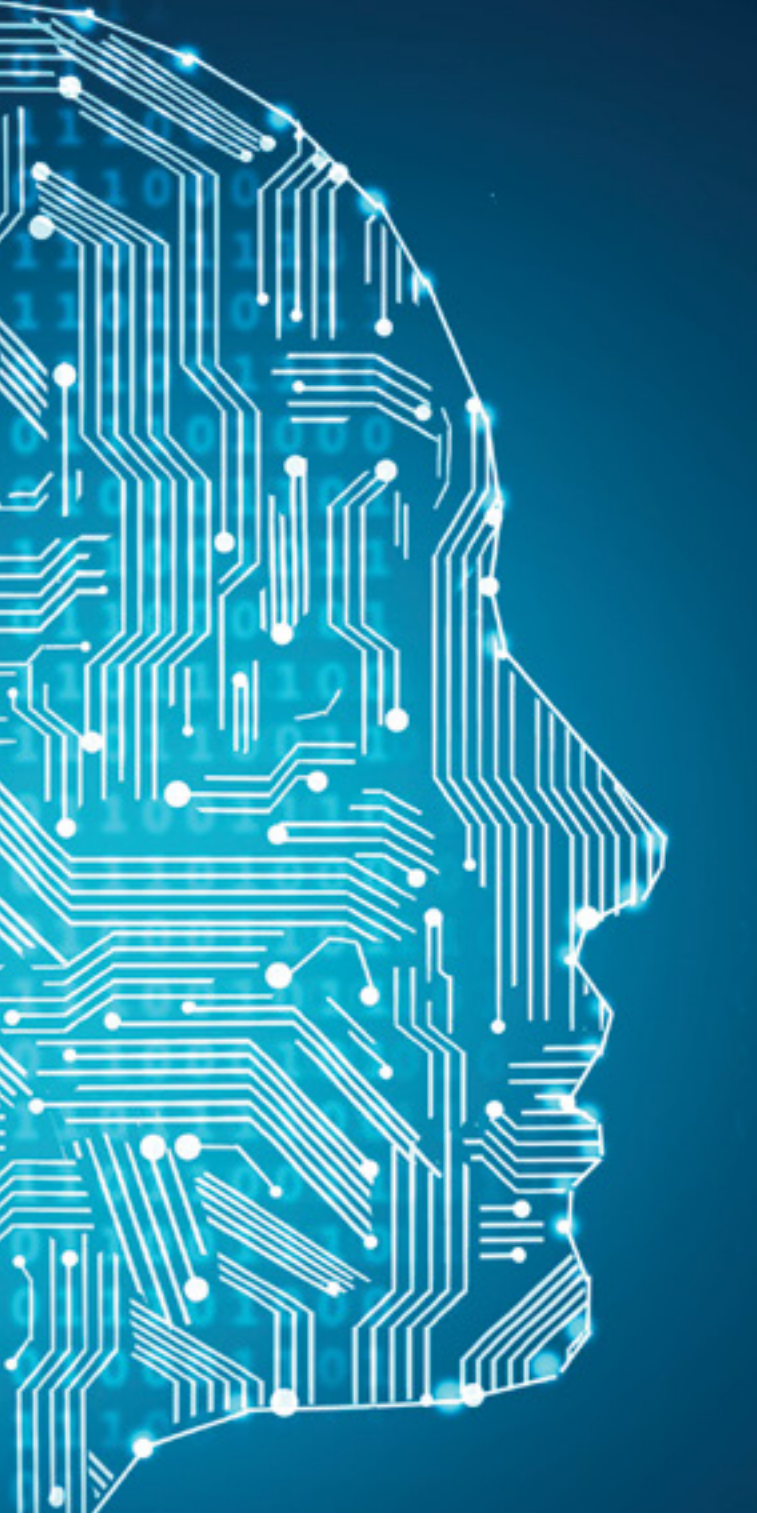

# AI FOR EXPLOSIVE ORDNANCE DETECTION IN CLEARANCE OPERATIONS: THE STATE OF RESEARCH

By Björn Kischelewski [ University College London ],
Gregory Cathcart [ Independent ],
David Wahl [ Scientia Education ], and
Benjamin Guedj [ University College London / Inria ]

The detection and clearance of explosive ordnance (EO) continues to be a predominantly manual and high-risk process that can benefit from advances in technology to improve its efficiency and effectiveness. Research on artificial intelligence (AI) for EO detection in clearance operations has grown significantly in recent years. However, this research spans a wide range of fields, making it difficult to gain a comprehensive understanding of current trends and developments. Therefore, this article provides a literature review of academic research on AI for EO detection in clearance operations. It finds that research can be grouped into two main streams: AI for EO object detection and AI for EO risk prediction, with the latter being much less studied than the former.

From the literature review, we develop three opportunities for future research. These include a call for renewed efforts in the use of AI for EO risk prediction, the combination of different AI systems and data sources, and novel approaches to improve EO risk prediction performance, such as pattern-based predictions.

Finally, we provide a perspective on the future of AI for EO detection in clearance operations. We emphasize the role of traditional machine learning (ML) for this task, the need to dynamically incorporate expert knowledge into the models, and the importance of effectively integrating AI systems with real-world operations.

## OVERVIEW OF RESEARCH ON AI FOR EXPLOSIVE ORDNANCE DETECTION IN CLEARANCE OPERATIONS

Survey and clearance remain predominantly manual and risky processes with increasingly limited financial resources. Technology has the potential to significantly improve this process. Thus, public and private organizations actively seek innovative research in technology that improves clearance efficiency and effectiveness.[1,2,3,4] For example, the Oslo Action Plan, a political commitment signed by States Parties implementing obligations of the *Anti-Personnel Mine Ban Convention*, explicitly states in action 27 to "[t]ake appropriate steps to improve the effectiveness and efficiency of survey and clearance, including by promoting the research, application and sharing of innovative technological means to this effect."[5]

Research on AI for mine action has seen a significant surge in recent years. At the Innovation Conference 2023, organized by the Geneva International Centre for Humanitarian Demining (GICHD), AI and data analysis were featured prominently, being the focus of five out of eleven breakout sessions.[6] However, this research spans a diverse range of fields, including electrical engineering, statistics, and more, making it challenging to gain a comprehensive understanding of current trends and developments. Nevertheless, such an overview is essential for both researchers and practitioners in mine action, as it helps identify promising research areas, ensures the efficient allocation of resources, and facilitates real-world impact.

To address this, we conducted a comprehensive review of research publications on AI systems for EO detection to support clearance operations, examining one hundred eligible publications. The earliest eligible publication is from 1997 (see Figure 1). Specifically, it explores the types of data and AI models utilized in the research to identify gaps and suggests directions for future work. Our literature review followed a multi-step process with several screening stages to identify relevant studies. An initial keyword search across seven databases yielded 1,558 results. These publications were then thoroughly screened based on predefined inclusion criteria. Only empirical, peer-reviewed studies using real-world data (excluding computer simulations) and applying ML algorithms for EO detection were included.

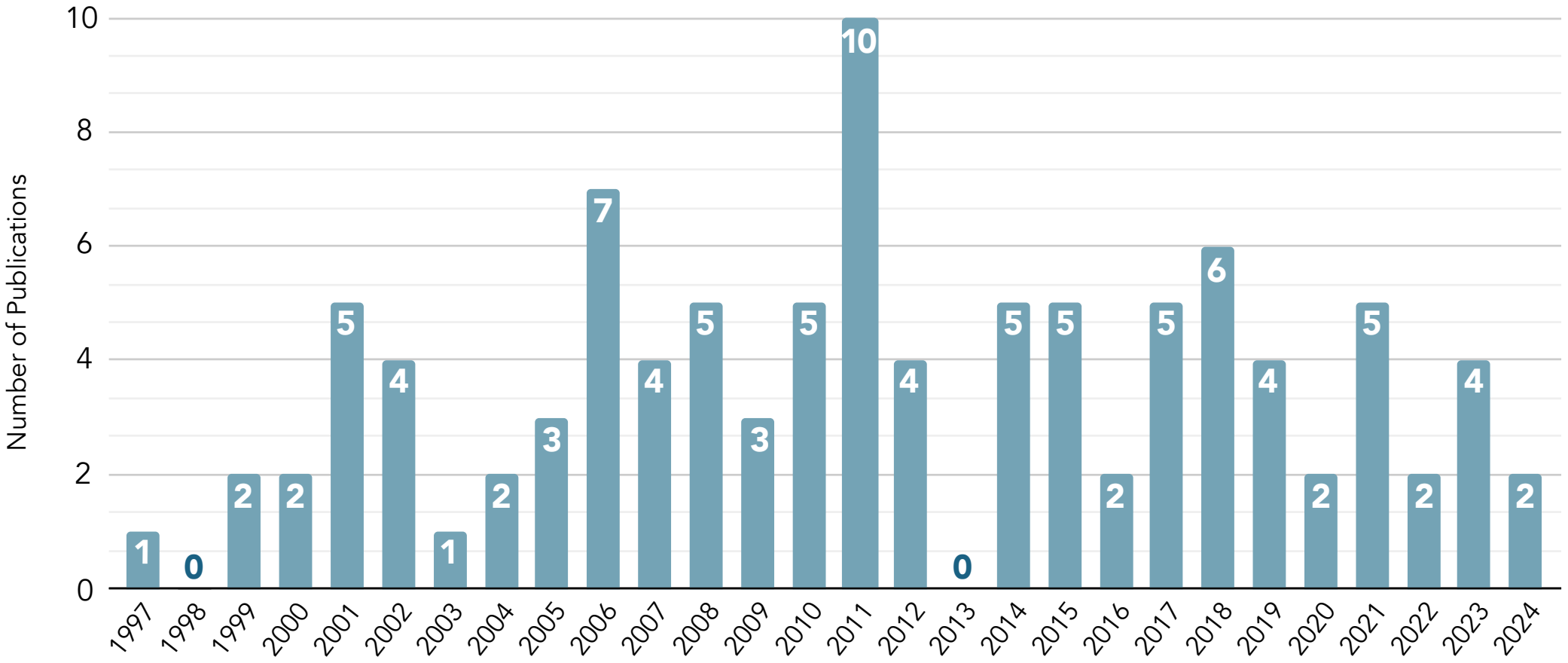

**Figure 1.** Number of publications per year on artificial intelligence systems for explosive ordnance detection in clearance operations.
*All graphics courtesy of the authors.*

## THE STATE OF RESEARCH COMPRISES TWO MAIN STREAMS

The review revealed that research on AI systems for EO detection in clearance operations can be grouped into two main streams (see Figure 2). More than 90 percent of the publications focus on EO object detection, including research on AI systems to localize EO during clearance operations. In contrast, a minority of the publications focus on EO risk prediction. Risk prediction seeks to recognize patterns in data (mine-laying patterns or patterns in geographical data) to determine the likelihood of finding EO in a region of interest to support survey and clearance operations. Both streams use different input data for the predictions.

Across both research streams, a variety of AI models are developed and evaluated. In total, thirty different types of AI algorithms are used (see Figure 3). Most of these algorithms can be divided into five groups, which are analyzed in at least ten different publications. These groups are (convolutional) neural networks (thirty-nine publications), support vector machines (twenty-seven publications), linear or logistic regression (twelve publications), hidden Markov models (eleven publications), and tree-based algorithms (ten publications). Understanding the AI approaches employed also helps identify potential leads for new research.

## RESEARCH STREAM 1: AI FOR EXPLOSIVE ORDNANCE OBJECT DETECTION

The majority of the research focuses on object detection. These studies cover a wide range of input data types that can be grouped into sensor data, including data from ground penetrating radar (GPR) and metal detectors, and image data, including thermal and hyperspectral image data. The researchers use AI to improve the detection of individual EO objects from the data streams.

Most sensor research uses GPR data, starting with Agarwal et al. in 2001.[7] Other authors apply AI techniques to various types of metal detectors, such as magnetometers or electromagnetic induction (EMI) sensors. They also use data from microwave sensors or from contact pressure sensors.[8,9] Most commonly, these studies use sensor data to train support vector machines (SVM) or artificial neural networks (ANN). They find that SVMs perform well in detecting different types of EO from GPR data.[10,11,12] However, Bray et al. find that an ANN outperforms an SVM for EO detection when trained and tested with data from a metal detector.[13] In general, authors find that ANNs for EO detection from metal detector data generalize well between different test sites.[14,15,16]

In contrast to sensor data, only sixteen studies use image data. Most authors use hyperspectral imagery, such as Bolton and Gader,[17] who use airborne imagery with seventy spectral responses. However, others such as Baur et al.,[18] use only RGB (red, green, blue) imagery, which they acquired by unmanned aerial vehicles (UAV). Another ten researchers use infrared or thermal image data to detect both buried and surface EO.[19,20] Most of these publications use convolutional neural networks (CNN). The authors report high performance with detection rates of up to 90 percent.[21,22] Pre-trained CNNs are also found to perform well in this task.[23,24] In addition, Thomas and Cathcart propose an extension to image-based algorithms by incorporating pattern information from EO to reduce the false alarm rate of predictions.[25]

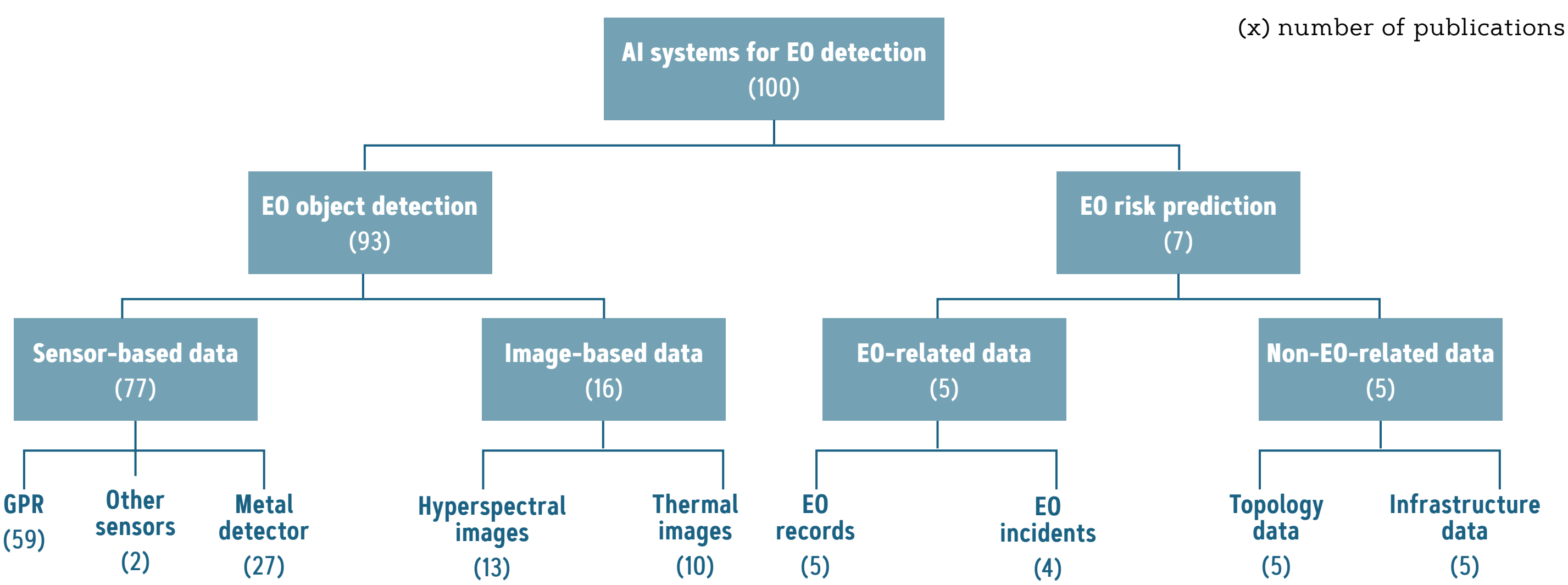

**Figure 2.** Research streams on artificial intelligence systems for explosive ordnance detection in clearance operations.

Furthermore, AI-based sensor fusion for EO object detection is analyzed by fourteen authors. Most of them combine GPR and metal detector data to improve EO object detection results.[26] However, two publications combine EMI and magnetometer data to improve detection performance.[27,28] Two other publications focus on sensor fusion of different image sensors such as RGB and thermal image data.[29,30]

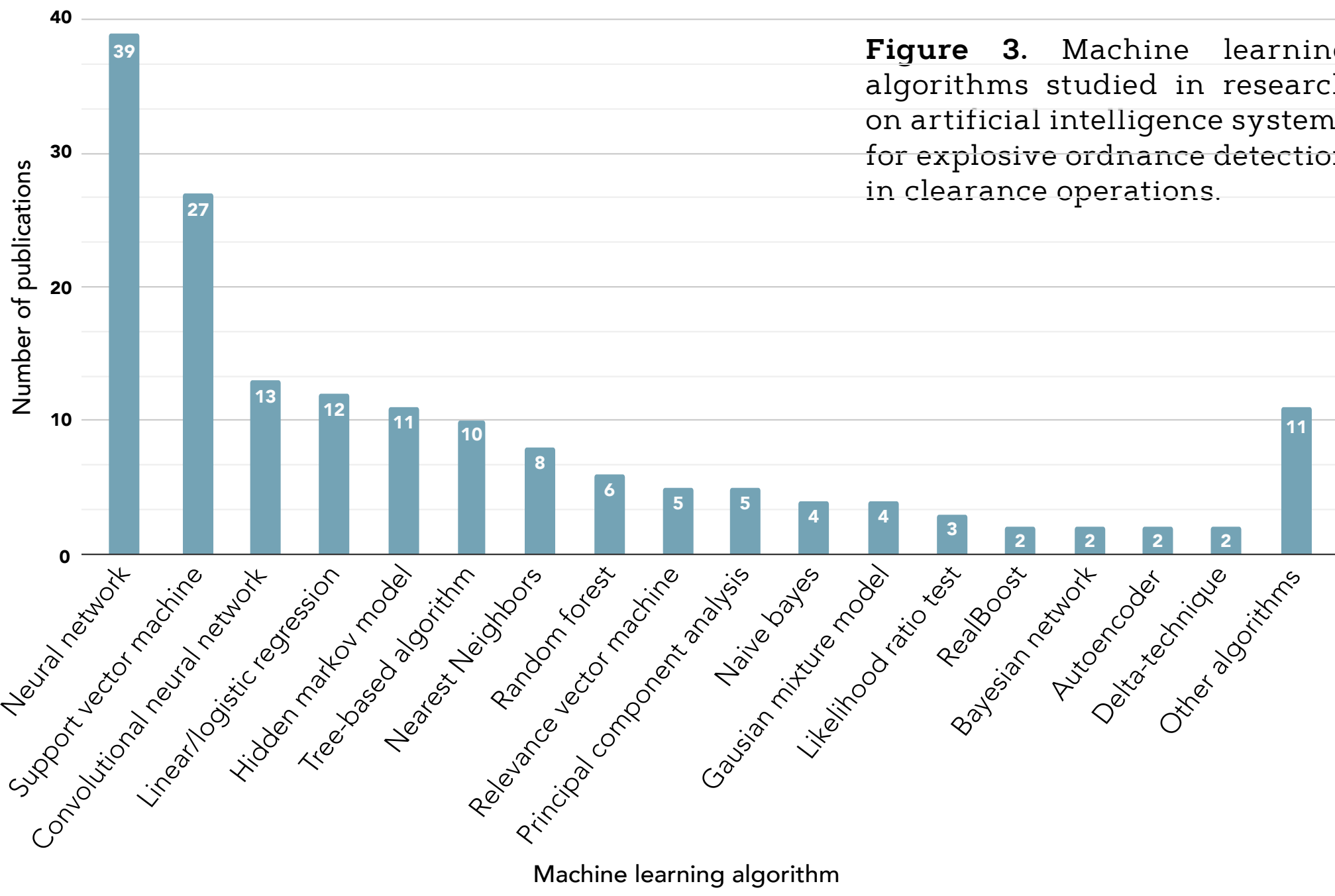

**Figure 3.** Machine learning algorithms studied in research on artificial intelligence systems for explosive ordnance detection in clearance operations.

## RESEARCH STREAM 2: AI FOR EXPLOSIVE ORDNANCE RISK PREDICTION

The second stream of research focuses on EO risk prediction for a region of interest and includes seven publications. These authors use input data that can be divided into EO-related and non-EO-related data. The former includes data from EO records as well as from EO incidents in the region of interest. This data is used to calculate input features for the risk prediction, such as distance to the nearest EO incident or to the front line. For example, Alegria et al. use landmine incident data to train their model.[31] Riese et al. also use several features related to EO locations, such as the distance to the nearest minefield, the distance to the confrontation line, and the distance to the nearest recorded mine accident to train their model.[32] The non-EO-related data includes various information on the topology and infrastructure of the region of interest. Topological features contain information on elevation, incline, land use, forests, rivers, animal density, soil texture, temperature, rainfall, and visibility. In addition, infrastructure features contain information about roads, railways, airfields, seaports, bridges, cities, buildings, financial institutions, schools, borders, telecommunication lines, power lines, oil lines, orchards, bunkers, trenches, and shelters.[33,34,35,36,37] Interestingly, most publications on EO risk prediction use a combination of EO-related and non-EO-related data.

All of these publications focus mainly on fundamental AI algorithms such as SVM, tree-based, and nearest neighbor models. For example, Rafique et al. compare a logistic regression model with an SVM. They find that the SVM outperforms the logistic regression and generalizes better across different regions.[38] Further, Saliba et al. compare an SVM with a random forest and an XGBoost model and find that the random forest model performs best.[39] The other publications in this research stream analyze neural networks, tree-based models, nearest neighbors models, principal component analysis, and Naive Bayes approaches. However, they all focus on a single ML technique and therefore cannot provide a meaningful performance comparison.[40,41,42,43,44]

## THREE OPPORTUNITIES FOR FUTURE RESEARCH

The analysis of the eligible publications reveals several gaps in current research on AI for EO detection in clearance operations, which translate into three suggested streams for future research:

1. **Renewed effort on using AI for EO risk prediction.** New research could make significant contributions to the ability of AI to predict the risk of finding EO at a given location, for example by identifying the most relevant input features for EO risk prediction. In addition, research could continue to compare different ML techniques for prediction, as this has only been studied by two publications.[45,46] Identifying the best performing ML techniques with improved detection rates will create real world impact in the field.
2. **Exploration of the combination of different AI systems.** From an AI perspective, both the object detection and risk prediction streams produce an estimate of the probability of finding EO at a given location. Thus, predictions from AI systems using airborne imagery could be used as input to GPR sensor data to reduce false alarm rates. Additionally, such systems could be improved by integrating results from AI-based EO risk predictors that use topology and EO-related data. For example, such risk predictions could be used to cross-check image data for false negatives. Finally, all AI systems could incorporate mine action expertise from the field, for instance by incorporating prior knowledge into the prediction or by providing explanations of the predictions.
3. **Novel approaches to improve the performance of EO risk prediction.** Namely, future research should investigate the extension of algorithms with EO pattern information for object detection and risk prediction, as proposed by Thomas and Cathcart.[47] The authors show improvements in EO detection by learning simple grid patterns. This approach could be extended in future research to more complex patterns and applied to EO object detection and risk prediction.

## THE FUTURE OF AI FOR EXPLOSIVE ORDNANCE DETECTION IN CLEARANCE OPERATIONS

Despite the potential of AI to enhance the efficiency and effectiveness of EO detection in clearance operations, some researchers criticize recent efforts as "potentially hazardous, insufficiently tested, and unlikely to provide practical solutions."[48] Thus, leveraging the findings of this literature review to develop a fact-based perspective on the future of AI for EO detection in clearance operations is crucial.

The rapid evolution of AI algorithms, from traditional ML to deep learning techniques, including large language models (LLM), means that there is a wide range of AI techniques that can be applied to EO detection in clearance operations. Although deep learning techniques are often used, many researchers report better or equal results from less complex ML techniques, which also allow for greater explainability. In the future, AI research for EO detection in clearance operations may split into different threads. While some will focus on applying the latest advances in AI, **traditional ML techniques will likely continue to play the primary role** in this field. Recent advances in generative AI and LLMs do not fit the problem of EO detection in clearance operations. Thus, concerns about their risks, such as those raised by Gasser,[49] will have little relevance to using AI for EO detection in clearance operations.

As in many fields where AI is used to enhance and reduce the risk of human activities, **AI is likely to work hand in hand with field experts** to improve their work, as anticipated by the GICHD.[50] Fully automated EO clearance through AI remains a distant prospect and, as this literature review shows, is unlikely to be the immediate focus of research efforts. Therefore, AI can be viewed as a useful tool for experts in the field rather than a solution to the challenges of EO clearance, thereby limiting the risks and responsibilities of the system.

While AI offers significant potential for EO detection in clearance operations, the sector must **ensure that it is effectively integrated into real-world operations**. This requires training and testing models under realistic conditions with standardized test data to ensure comparability. Further, it requires conducting usability research to understand the needs of human operators and incorporating prior expert knowledge to improve model robustness. In this regard, we emphasize the importance of developing international funding streams that can drive collaboration between states, research institutions, and mine action programs to ensure that innovation is intrinsically linked to the specific nature and challenges of removing contamination, reducing risk, and improving lives in regions affected by EO.

As the technology continues to advance, it is also important to remain vigilant and ensure that the development and deployment of AI is guided by ethical principles and a commitment to humanitarian goals. By addressing these challenges, AI can become a powerful tool for EO detection in clearance operations, helping to save lives and accelerate the process of clearing contaminated areas.

**BJÖRN KISCHELEWSKI**
**Consultant for Cloud, Data, and Artificial Intelligence**
**University College London**

Björn Kischelewski is a consultant for cloud, data and artificial intelligence (AI), and has studied AI for Sustainable Development at University College London where he conducted research on AI systems for mine action. In particular, he explored the use of landmine pattern information for landmine risk prediction to improve clearance efficiency.

**DAVID WAHL**
**Lead Researcher**
**Scientia Education**

David Wahl is the lead researcher in application of technology to solve humanitarian problems at Scientia Education. He has previously worked as a quantitative researcher for a large hedge fund in London and lectured in physics in South America.

**GREGORY CATHCART**
**Implementation Support Officer**
**Anti-Personnel Mine Ban Convention Implementation Support Unit**
**Geneva International Centre for Humanitarian Demining**

Gregory Cathcart advises countries, UN Agencies, international nongovernmental agencies,and CSOs on matters related to the implementation of mine action. Cathcart holds a Master of Public Health from Flinders University, Australia. He contributed the chapter, "Landmines as a form of community protection in Eastern Myanmar" in the Australian National University's publication, *Conflict in Myanmar: War, Politics, Religion.*

**BENJAMIN GUEDJ**
**Professor of Machine Learning and Foundational Artificial Intelligence, University College London**
**Research Director, Inria**
**Turing Fellow, The Alan Turing Institute**

Benjamin Guedj is a senior scientist conducting research in machine learning (ML) and AI in the United Kingdom and France. He is a Professor of Machine Learning and Foundational Artificial Intelligence with University College London (Centre for AI, Department of Computer Science) and a Research Director with Inria. He is also a Turing Fellow with the Alan Turing Institute, and co-director of the ELLIS London Unit. His interests lie in the foundations of AI and theory of ML, and he has contributed to the understanding of intelligent systems through statistical learning theory and the PAC-Bayes theory. Guedj serves on the scientific committees of many machine learning venues and he is a Knight of the Order of the Academic Palms of the French Republic.

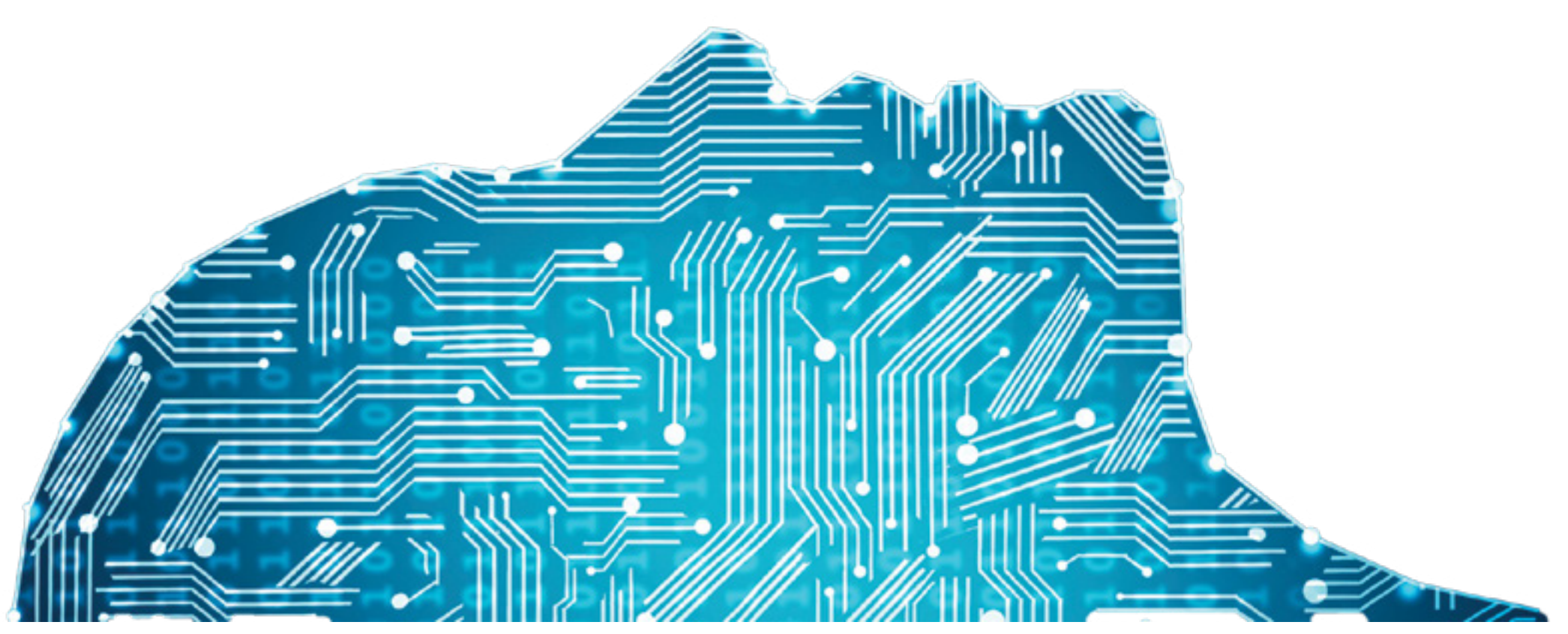